\documentclass[runningheads]{llncs}
\usepackage{graphicx}
\usepackage{amsmath,amssymb} % define this before the line numbering.
\usepackage{color}
\def\koncept{Koncept512}
\def\frugan{EvolGAN}
\def\ava{AVA}
 
\makeatletter
\newcommand{\printfnsymbol}[1]{%
  \textsuperscript{\@fnsymbol{#1}}%
}
\makeatother
\usepackage{booktabs} % For formal tables
\usepackage{url,rotating,ifthen,epsfig,multirow,array,color,multicol,hhline,calc,algpseudocode}
\usepackage{todonotes} 
\usepackage{bold-extra,ulem}
\usepackage{enumitem}
\setlistdepth{9}
\setlist[itemize,1]{label=$\bullet$}
\setlist[itemize,2]{label=$-$}
\setlist[itemize,3]{label=$\ast$}
\setlist[itemize,4]{label=$\cdot$}
\setlist[itemize,5]{label=$\diamond$}
\renewlist{itemize}{itemize}{5}

\newcommand{\otc}[1]{\textcolor{black}{#1}}
\newcommand{\olivieraccv}[1]{\textcolor{black}{#1}}
\newcommand{\baptisteaccv}[1]{\textcolor{black}{#1}}
\newcommand{\mariiaaccv}[1]{\textcolor{black}{#1}}
\newcommand{\fabienaaccv}[1]{\textcolor{black}{#1}}
\newcommand{\vladaccv}[1]{\textcolor{black}{#1}}

\newcommand{\ft}[1]{\textcolor{black}{#1}}
\newcommand{\br}[1]{\textcolor{black}{#1}}

\newcommand{\ccc}[1]{\textcolor{black}{#1}}

\newcommand{\brtwo}[1]{\textcolor{black}{#1}}
\newcommand{\brthree}[1]{\textcolor{black}{#1}}
\newcommand{\brcameraready}[1]{\textcolor{blue}{#1}}

 \usepackage[algoruled,linesnumbered]{algorithm2e}
\algdef{SE}[PROCEDURE]{Procedure}{EndProcedure}%
[2]{\algorithmicprocedure\ \textproc{#1}\ifthenelse{\equal{#2}{}}{}{(#2)}}%
{\algorithmicend\ \algorithmicprocedure}%
   
\algdef{SE}[FUNCTION]{Function}{EndFunction}%
[2]{\algorithmicfunction\ \textproc{#1}\ifthenelse{\equal{#2}{}}{}{(#2)}}%
{\algorithmicend\ \algorithmicfunction}%
\usepackage{hyperref,float,placeins}
\hypersetup{
    colorlinks,
    citecolor=black,
    filecolor=black,
    linkcolor=blue,
    urlcolor=black}

\def\E{{\mathbb E}}

\def\PP{{\cal P}}

\def\R{{\mathbb R}}
\begin{document}
\pagestyle{headings}
\mainmatter
\def\ACCV20SubNumber{940}  % Insert your submission number here

\sloppy
%\begin{document}

\def\finalversion{
\title{\frugan{}: Evolutionary Generative Adversarial Networks}
\renewcommand{\topfraction}{2.95}
%\title{Test-based population-size adaptation: the noisy and the multimodal cases}
\author{Baptiste Roziere\inst{1}\thanks{Equal contribution} 
\and
Fabien Teytaud\printfnsymbol{1}\inst{2}
\and
Vlad Hosu\inst{3}
\and
Hanhe Lin\inst{3}
\and\\
Jeremy Rapin\inst{1}
\and
Mariia Zameshina\inst{4}
%\and
%Camille Couprie\inst{1} %C'est tres gentil Olivier mais dans les remerciements c'est tres bien pour ce que j'ai fait!
\and
Olivier Teytaud\inst{1}}
\authorrunning{B. Roziere et al.}
\institute{Facebook AI Research
\email{\{broz,jrapin,oteytaud\}@fb.com}\\
\and
Univ. Littoral Cote d'Opale \email{teytaud@univ-littoral.fr}
\and
Univ. Konstanz, Germany \email{\{hanhe.lin,vlad.hosu\}@uni-konstanz.de}
\and
Univ. Grenoble Alpes, CNRS, Inria, Grenoble INP, LIG, France
}
}

\finalversion{}
% \title{\frugan{}: Evolutionary Generative Adversarial Networks} % Replace with your title
% \titlerunning{\frugan{}: Evolutionary Generative Adversarial Networks}
% \authorrunning{ACCV-20 submission ID \ACCV20SubNumber}

% \author{Anonymous ACCV 2020 submission}
% \institute{Paper ID \ACCV20SubNumber}

% {
%Baptiste Roziere <-- Pytorch Gan Zoo xps,
%Fabien Teytaud <-- styleGan2 xps,
%Jeremy Rapin <-- Nevergrad
%Olivier Teytaud <-- styleGan2 xps, PokeGAN xps, latex
%
%* coded specifically for that project
%+ add KONCEPT512 guys
% }
\maketitle
\begin{abstract}
\baptisteaccv{We propose to use a quality estimator and evolutionary methods to search the latent space of generative adversarial networks trained on \vladaccv{small, difficult datasets, or both}. The new method leads to the generation of significantly higher quality images while preserving the \vladaccv{original generator's diversity}. Human raters preferred an image from the new version with frequency 83.7\% for Cats, 74\% for FashionGen, 70.4\% for Horses, and 69.2\% for Artworks - minor improvements for the already excellent GANs for faces. This approach \vladaccv{applies to} any quality scorer and GAN generator.}
% \olivieraccv{Various papers have proposed to search the latent space using various criteria: handcrafted criteria, simulation studies, neural classifiers providing class constraints. We show in the present document that using a quality estimator for biasing the technical quality of an image provides a significant improvement of the performance of an image generative adversarial network.}
\end{abstract}
\section{Introduction}    
Generative adversarial networks (GAN) \vladaccv{are the state-of-the-art generative models in many domains.}
% represent the state of the art for generative modeling in many domains. %, in particular in computer vision.
However, they need quite a lot of training data \vladaccv{to reach} a decent performance.
Using off-the-shelf image quality estimators, we propose a novel but simple evolutionary modification for making them more reliable for small, difficult, or multimodal datasets. Contrarily to previous approaches \brthree{using evolutionary methods for image generation, we do not modify the training phase}. We use a generator $G$ mapping a latent vector $z$ to an image $G(z)$ built as in a classical GAN. The difference lies in the method used for choosing a latent vector $z$. Instead of randomly generating a latent vector $z$, we perform an evolutionary optimization, with $z$ as decision variables and the estimated quality of $G(z)$~\brtwo{\textemdash}~based on a state-of-the-art quality estimation method\brtwo{\textemdash} as an objective function. We show that:
\begin{itemize}
    \item The quality of generated images is better, both for the proxy used for estimating the quality, i.e., the objective function, as well as for human raters. For example, %Tables \ref{kanimals} and \ref{xptable}} show that 
    the modified images are \baptisteaccv{preferred} by human raters more than 80\% of the time  for images of cats and around 70\% of the time for horses and artworks.
    \item The diversity of the original GAN is preserved: % (Fig. \ref{pairing} and \ref{pairingpgan}\otc{: when we use our modified variant with $\alpha<<1$ (Sec. \ref{local}), 
    the new image\ft{s are} \baptisteaccv{preferred by humans and still similar}.
    \item The computational overhead \ft{introduced by the evolutionary optimization} is moderate, compared to the computational requirement for training the original GAN.
\end{itemize}
The approach is simple, generic, easy to implement \baptisteaccv{, and fast. It} can be used as a drop-in replacement for classical GAN provided that we have a quality estimator for the outputs of the GAN. Besides the training of the original GAN, many experiments were performed on a laptop without any GPU.
\begin{figure}
    \centering
\begin{tabular}{cc@{\vladaccv{\hskip 10px}}cc}
\multicolumn{2}{c}{\vladaccv{\textbf{StyleGAN2}}} & \multicolumn{2}{c}{\vladaccv{\textbf{EvolGAN}}} \\
% \baptisteaccv{\textbf{StyleGAN2}}& \baptisteaccv{\textbf{StyleGAN2}}& \baptisteaccv{\textbf{EvolGAN}}& \baptisteaccv{\textbf{EvolGAN}}\\
\includegraphics[width=.24\textwidth]{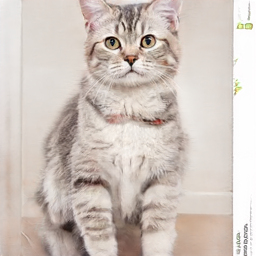}&
\includegraphics[width=.24\textwidth]{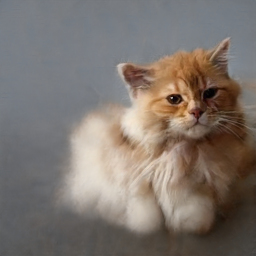}&
\includegraphics[width=.24\textwidth]{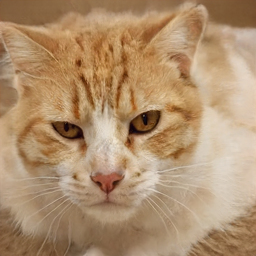}&
\includegraphics[width=.24\textwidth]{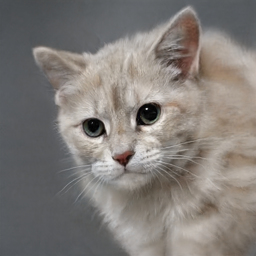}\\
\addlinespace[7px]
\includegraphics[width=.24\textwidth]{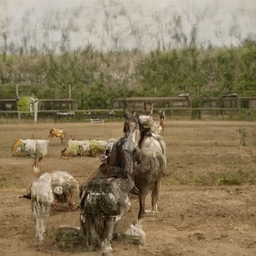}&
\includegraphics[width=.24\textwidth]{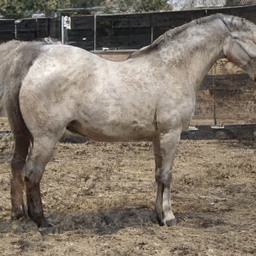}&
\includegraphics[width=.24\textwidth]{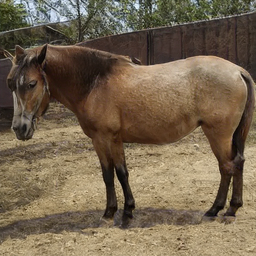}&
\includegraphics[width=.24\textwidth]{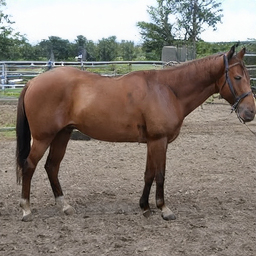}\\
\includegraphics[width=.24\textwidth]{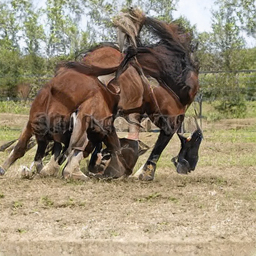}&
\includegraphics[width=.24\textwidth]{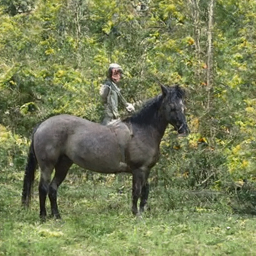} &   
\includegraphics[width=.24\textwidth]{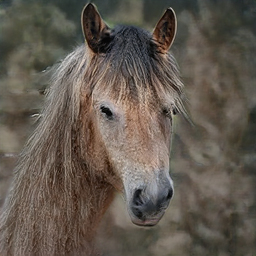}&
\includegraphics[width=.24\textwidth]{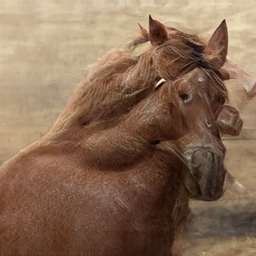}\\
\end{tabular}
\caption{\baptisteaccv{For illustration , random images generated using StyleGAN2 (left) and \frugan{} (right). Horses were typically harder than cats. The images generated by \frugan{} are \vladaccv{generally} more realistic.} \vladaccv{The top-left example of a generated cat by } StyleGAN2 has \brthree{blood-like} artifacts on \brthree{its} throat and the other is blurry. Three of the four StyleGAN2 horses \brtwo{are clearly unrealistic}: \mariiaaccv{on the bottom right} \olivieraccv{of the StyleGan2 results,} the human and the horse are mixed, the bottom left shows \vladaccv{an incoherent} mix of \vladaccv{several} horses, % \vladaccv{} with legs and tails everywhere; 
the top left looks like the ghost, and only the top right is \baptisteaccv{realistic}. Overall, \vladaccv{both} cats, and 3 of the 4 horses \vladaccv{generated} by \frugan{} look realistic. \vladaccv{We show more examples of horses, as they are more difficult to model.}}
    % \caption{First and second columns: images generated by StyleGAN2. Third and fourth column: images generated by \frugan{} parametrized as \frugan{}($b=40$, $\alpha=\infty$). One of the StyleGAN2 cats has \brthree{blood-like} artifacts on \brthree{its} throat and the other is blurry, but they are \brthree{quite realistic} overall. Three of the four StyleGAN2 horses \brtwo{are clearly unrealistic} (\mariiaaccv{on the bottom right} the human and the horse are mixed; the bottom left horse looks like a mix of 7 horses with legs and tails everywhere; the top left looks like the ghost of a horse; only the top right looks good. The 2 cats and 3 of the 4 horses by \frugan{} look realistic.}
    \label{samples}
\end{figure}
%\setcounter{tocdepth}{5}
%\tableofcontents
    %\todo{source code ?}\todo{more pokeGan XPs with less budget ?} \todo{robustness indicators, like which methods provide the worst ?}\todo{discuss mutation rates papers ?}
    %\todo{Specify datasets; see if ``Frugal'' is justified; check datasets size for traditional GAN: CelebaHQ: 30000, FashionGen: 293000 but not sure it's the same thing in PytorchGanZoo}
\begin{figure}[t]
    \centering
    %\begin{minipage}{.58\textwidth}
        \includegraphics[trim = 90 200 70 0, clip, width=0.53\textwidth]{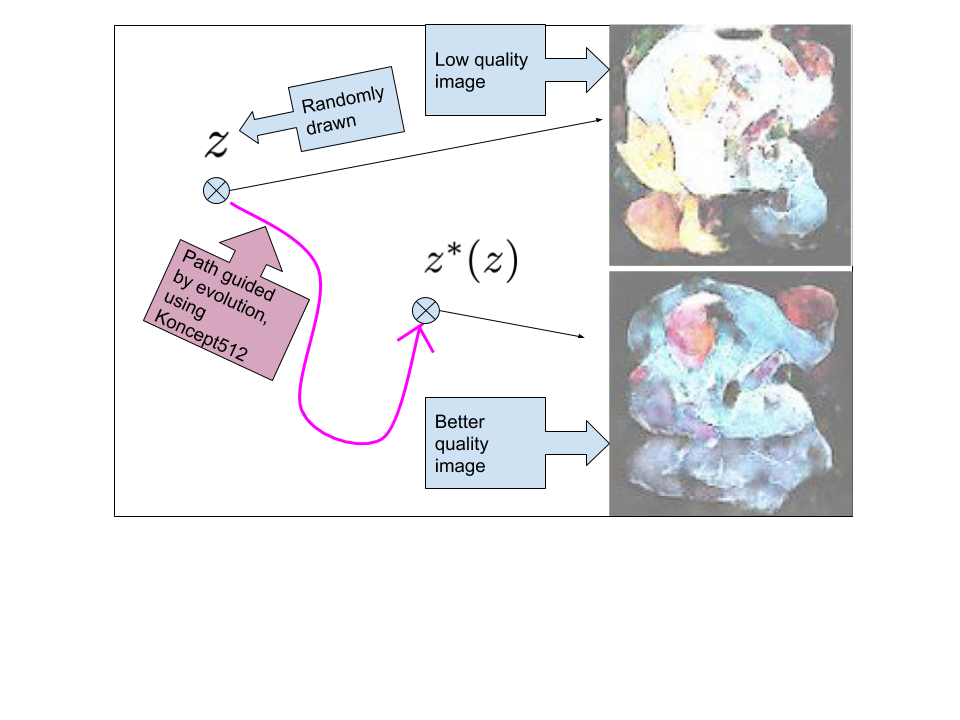}
          \includegraphics[width=0.43\textwidth]{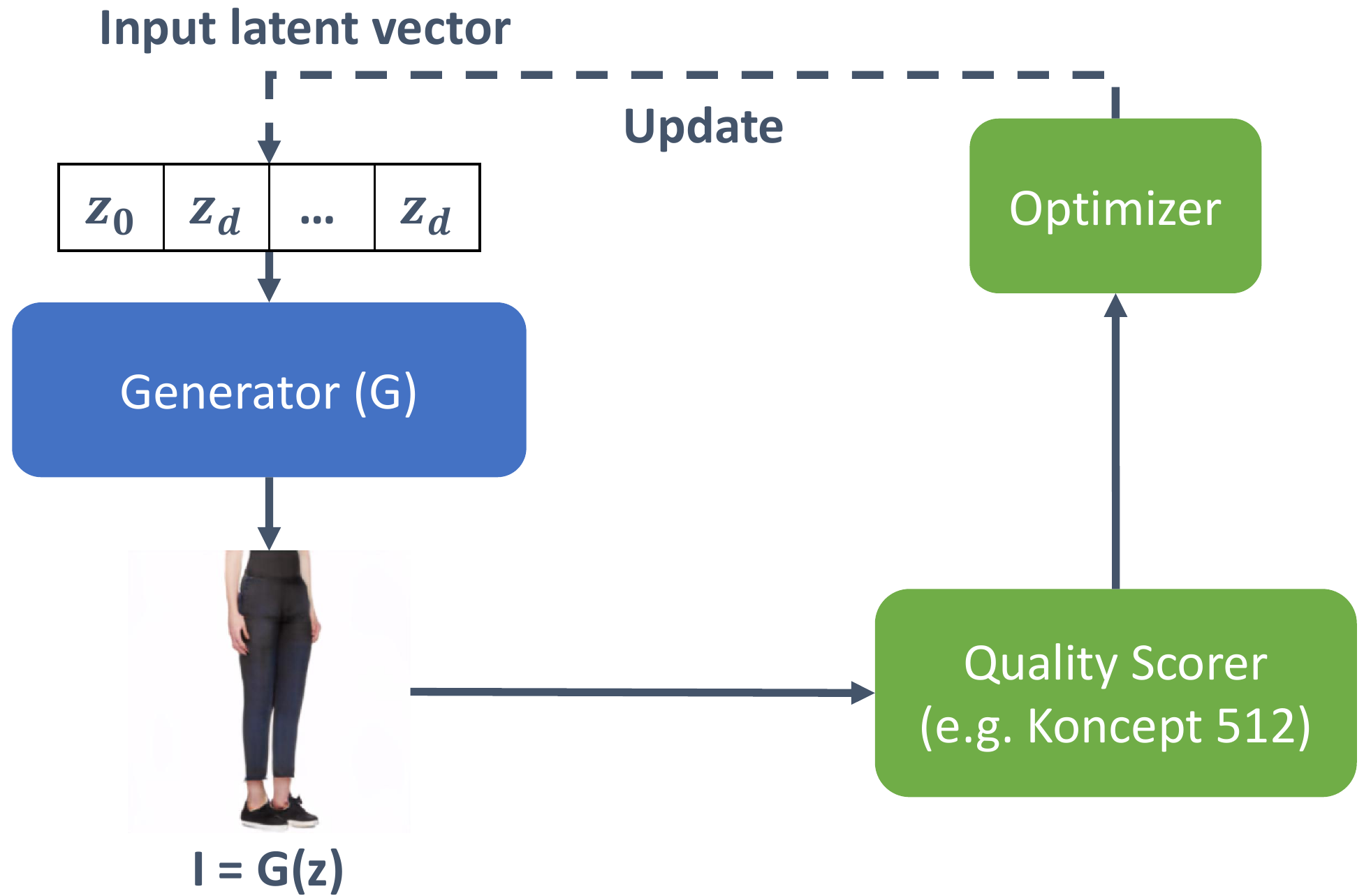}\\
    %\end{minipage}
    %\begin{minipage}{.35\textwidth}
    % We keep the same generator \otc{$G$ mapping some latent vector $z$ to an image $G(z)$ ($G$ is }typically created by a generative adversarial network) but modify the initial $z$: whereas classical models randomly choose $z$, we modify that randomly chosen $z$ into a better $z^*(z)$ evolved for improving the quality as estimated by, typically, \koncept{}.
    %\end{minipage}
    \caption{General \frugan{} approach (left) and optimization loop (right).
    \br{Our method improves upon a pre-trained generator model G, which maps a latent vector $z$ to an image G(\ft{$z$}). In classical models, the images are generated by sampling a latent vector $z$ randomly.} We modify that randomly chosen $z$ into a better $z^*(z_0)$ evolved \br{to improve the quality of the image, typically estimated by} \koncept{}.
    \otc{For preserving diversity, we ensure that $z^*(z_0)$ is close to the original $z_0$.} The original image is generated using PokeGan (Pokemon dataset) and the improved one is generated using \frugan{} (superposed on top of PokeGan): we see an elephant-style Pokemon, hardly visible in the top version.}
    \label{fig:frugan}
\end{figure}
Fig. \ref{samples} shows examples of generations of $\frugan{}_{StyleGAN2}$ compared to generations by StyleGAN2. \olivieraccv{Fig. \ref{fig:frugan} presents} our general approach, detailed in the \olivieraccv{Section \ref{sec:method}}. 
\section{Related Works}
\subsection{Generative Adversarial Networks}
\olivieraccv{
Generative Adversarial Networks \cite{goodfellow2014generative} (GANs) are widely used in machine learning
\cite{sbai2018design,ZhuUrtasun2017iccvPrada,elgammal17,Zhu2017cycleGANs,Park2019spade,Donahue2016Bigan,frid2018gan,nie2017medical} for generative modeling.
\baptisteaccv{Generative Adversarial Networks are made up of two neural networks: a Generator $G$, mapping a latent vector $z$ to an image $G(z)$ and a Discriminator $D$ mapping an image $I$ to a realism value $D(I)$.
Given a dataset ${\cal D}$, \fabienaaccv{GANs} are trained using two training steps operating concurrently:}
\begin{enumerate}[label=(\roman*)]
            \item Given a randomly generated $z$, \br{the generator $G$ tries to fool $D$ into classifying its output $G(z)$ as a real image, e.g. by maximizing $\log D(G(z))$. For this part of the training, only the weights of G are modified.} %to maximize $D(G(z))$, so that its outputs $G(z)$ is, according to the discriminator $D$, realistic. This means that we maximize $D(G(z))$, with as variables the weights of $G$. For this part of the training, the weights of $D$ are not modified.
            \item Given a minibatch containing both random fake images $F=\{G(z_1),\dots,G(z_k)\}$ and real images $R=\{I_1,\dots,I_k\}$ randomly drawn in ${\cal D}$, the discriminator learns to distinguish $R$ and $F$ \br{, e.g. by optimizing the cross-entropy}.
\end{enumerate}
%Their ability
\baptisteaccv{The} \mariiaaccv{ability of GANs} to \vladaccv{synthesize} faces~\cite{stylegan2} \baptisteaccv{is particularly impressive \vladaccv{and of wide interest}}. However, \br{such results are} possible only with huge datasets for each modality and/or after careful cropping, which restricts their applicability. }
\vladaccv{Here we} consider \otc{the problem of improving GANs trained on small or difficult datasets.}
\olivieraccv{Classical tools for making GANs compatible with small datasets include:
\begin{itemize}
    \item Data augmentation, by translation, rotation, symmetries, or other transformations.
    \item Transfer from an existing GAN trained on another dataset to a new dataset \cite{transfergan}.
    \item %Several papers modify the distribution for matching a specific request. 
    \mariiaaccv{Modification of the distribution in order to match a specific request as done in several papers.}
    \cite{parimala2019quality} modifies the training, using quality assessement as we do; however they modify the training whereas we modify inference. In the same vein, \cite{yi2020bsd} works on scale disentanglement: it also works at training time. These works could actually be combined with ours.  
    \cite{rr1} generates images conditionally to a classifier output or conditionally to a captioning network output. \cite{rr2} and \cite{rr3} \fabienaaccv{condition} the generation to a playability criterion (estimated by an agent using the GAN output) or some high-level constraints. 
    \cite{rr4} uses a variational autoencoder (VAE), so that constraints \fabienaaccv{can be added to the generation}: they can add an attribute (e.g. black hair) and still take into account a realism criterion extracted from the VAE: this uses labels from the dataset. \cite{rr5} uses disentanglement of the latent space for semantic face editing: the user can modify a specific latent variable.
    \cite{zhu2016generative} allows image editing and manipulation: it uses projections onto the output domain.
    \item \olivieraccv{Biasing the dataset. \baptisteaccv{\cite{mariani2018bagan} augments the dataset by generating images with a distribution skewed towards the minority classes.}}
\item Learning a specific probability distribution, rather than using a predefined, for example Gaussian, distribution. Such a method is advocated in \cite{deligan}.
    \end{itemize}
The latter is the closest to the present work in the sense that we stay close to the goal of the original GAN, i.e. modeling some outputs without trying to bias the construction towards some subset.}
However, whereas \cite{deligan} learn a probability distribution on the fly while training the GAN, our approach learns a classical GAN and modifies, a posteriori, the probability distribution by considering a subdomain of the space of the latent variables in which images have better quality. \otc{We could work on an arbitrary generative model \olivieraccv{based on latent variables}, not only GANs. As opposed to all previously mentioned works, we improve the generation, without modifying the target distribution and without using any \vladaccv{side-information} or handcrafted criterion - our ingredient is a quality estimator.}
Other combinations of deep learning and evolutionary algorithms have been published around GANs\br{. For instance,} \cite{otherevogan} evolves a population of generators, whereas our evolutionary algorithm evolves individuals in the latent space. \otc{\cite{tog} also evolves individuals in the latent space, but using human feedback rather than the quality estimators that we are using. \cite{camilleinspir} evolves individuals in the latent space, but either guided by human feedback or by using similarity to a target image.}
\subsection{Quality estimators: \koncept{} and \ava{}}
Quality estimation is a long-standing research topic \cite{wang2004image,ye2012unsup} recently improved by deep learning \cite{koncept512reference}. In the present work, we focus on such quality estimation tools based on supervised convolutional networks.
%\subsubsection{\koncept{}: technical quality estimator}
The KonIQ-10k dataset is a large publicly available image quality assessment dataset with 10,073 images rated by humans. Each image is annotated by 120 human raters.
\baptisteaccv{The \koncept{} image quality scorer}~\cite{koncept512reference} is based on an InceptionResNet-v2 architecture \br{and} trained \br{on KonIQ-10k} for predicting the mean opinion scores of the annotators.
\br{It} takes as input an image $I$ and outputs \br{a quality estimate} \baptisteaccv{$K(I)\in \R$}.
\koncept{} is the state of the art in technical quality estimation \cite{koncept512reference}, and is freely available. We use the release without any modification.
%\subsubsection{\ava{}: aesthetic quality estimator}
\cite{ava} provides a tool similar to \koncept{}, termed \ava{}, but dedicated to aesthetics rather than technical quality. It was easy to apply it as a drop-in replacement of \koncept{} in our experiments.
\section{Methods}\label{sec:method}
\subsection{Our algorithm: \frugan{}}\label{zefrugan}
%\begin{figure}[t]
%    \centering
   %   \caption{\frugan{}. Instead of returning $I = G(z)$ directly, we modify $z$ using a quality scorer and an optimizer.}
 %   \label{fig:frugan2}
%\end{figure}
We do not modify the training of the GAN. We use a generator $G$ created by a GAN. $G$ takes as input a random latent vector $z$, and outputs an image $G(z)$.
While the latent vector is generally chosen randomly (e.g., \br{$z\leftarrow \mathcal{N}(0,I_d)$}), we treat it as a free parameter to be optimized according to a quality criterion \baptisteaccv{$Q$}. 
More formally, we obtain \baptisteaccv{$z^*(z_0)$:
\begin{equation}
    z^*(z_0) = \arg\max_z Q(G(z))\mbox{ in the neighborhood of a random $z_0$.}\label{optim}
\end{equation}%\todo{Not sure we want to keep this equation. We optimize in the direction of $z^*$ but we don't actually get to a $z^*$ which is approximately equal to this (otherwise there would be no diversity}
In this paper, $Q$ is either \koncept{} or \ava{}.}
\baptisteaccv{Our algorithm computes an approximate solution of problem~\ref{optim} and outputs $G(z^*(z_0))$.}
Importantly, we do not want a global optimum of Eq. \ref{optim}. We want a local optimum, in order to have essentially the same image \baptisteaccv{-- $z^*(z_0)$} must be close to \baptisteaccv{$z_0$}, which would not happen without this condition.
The optimization algorithm used to obtain $z^*$ in \ft{Eq.} \ref{optim} is a simple $(1+1)$-Evolution Strategy with random mutation rates \cite{danglehre}, adapted as detailed in Section \ref{local} (see Alg. \ref{alg:frugan}). \br{We keep the budget of our algorithm low, and the mutation strength parameter $\alpha$ can be used to ensure that the image generated by \frugan{} is similar to the initial image. For instance, with $\alpha = 0$, the expected number of mutated variables is, by construction (see Section \ref{zefrugan}), bounded by $b$.}
We sometimes use \baptisteaccv{the aesthetic quality estimator} \ava{} rather than \baptisteaccv{the technical quality estimator} \koncept{} for quality estimation. We consider a coordinate-wise mutation rate\baptisteaccv{: we} mutate or do not mutate each coordinate, independently with some probability.
%\subsection{GANs}
\subsection{Optimization algorithms}\label{local}
After a few preliminary trials we decided to use the \fabienaaccv{$(1+1)$-Evolution Strategy} with uniform mixing of mutation rates~\cite{danglehre}, with a modification as described in algorithm \ref{alg:frugan}. \fabienaaccv{This modification is} designed for tuning the compromise between quality and diversity as discussed in Table \ref{tab}.
% {\bf{$\frugan{}_{G,b,\alpha}$:}}
% \todo{Change the presentation ( algorithm environment, maybe a table for the parameters) } Done
\begin{algorithm}[ht]
\SetKwInOut{Output}{Output}
\SetKwInOut{Parameters}{Parameters}
\Parameters{
    \begin{itemize}
    \item A probability distribution $\PP$ on $\R^d$.
    \item A quality estimator \baptisteaccv{$Q$}, providing an estimate \baptisteaccv{$Q(I)$} of the quality of some $I\in E$. We use $Q=\koncept$ or $Q=\ava$.
    \item A generator $G$, building $G(z)\in E$ for $z\in \R^d$.
    \item A budget $b$.
    \item A mutation strength $0\leq \alpha\leq \infty$.
    \item A randomly generated $z\leftarrow random(\PP)$. $I=G(z)$ is the baseline image we are trying to improve.
    \end{itemize}
    }
%$z':= z$ \\
\For{$i\in\{1,\dots,b\}$}{
$r := Clip(1/d, 1, \alpha\times uniform([0,1]))$\\ \label{alg:frugan:l2}
$z':= z$ \\
\For {$j\in \{1,\dots,d\}$}{
with probability $r$, $z'_i\leftarrow random(\PP)_i$ ($i^{th}$ marginal of $\PP$).
}
\baptisteaccv{\If {$Q(G(z)) < Q(G(z')$)}{ $z\leftarrow z'$}}
}
\Output{
Optimized image $I' = G(z)$}
\caption{The $\frugan{}_{G,b,\alpha}$ algorithm \label{alg:frugan}}
\end{algorithm}
We \baptisteaccv{used} $Clip(a,b,c)=\max(a,\min(b,c))$. Optionally, \baptisteaccv{$z_0$} can be provided as an argument, leading to \baptisteaccv{$\frugan{}_{G,b,\alpha,z_0}$}.
The difference with the standard uniform mixing of mutation rates is that $\alpha\neq 1$. With $\alpha=0$, \baptisteaccv{the resulting image} $I'$ is close to \baptisteaccv{the original image} $I$, whereas with $\alpha=\infty$ the outcome $I'$ is not similar to $I$. \baptisteaccv{Choosing} $\alpha=1$ (or $\alpha=\frac12$, closely related to FastGA\cite{fastga}) leads to faster convergence rates \baptisteaccv{but also to less diversity} \fabienaaccv{(see Alg.\ref{alg:frugan}, line \ref{alg:frugan:l2})}. %but loses the locality
We will show that overall, $\alpha=0$ is the best choice for \frugan{}.
We therefore get algorithms as \baptisteaccv{presented} in Table \ref{tab}.
\begin{table}[t]
    \centering
    \begin{tabular}{ccc}
    \toprule
        $0\leq \alpha \leq \infty$ &  behavior of $\frugan{}_{G,b,\alpha}$ & $\E||z^*(z_0)-z||_0$ with budget $b$\\
    \midrule
    \baptisteaccv{$0 \le \alpha \le \frac 1 d$} & standard $(1+1)$ evol. alg. & \\ &  with mutation rate $r=\frac1d$. & $\leq b$\\
    \hline
    $\alpha=1$ & uniform mixing of mutation rates & \\ &\cite{danglehre} (also related to~\cite{fastga}). &  \\
    \hline
    $\alpha=\infty$ & all variables mutated: equivalent & \\ & to random search & \\
    \hline
    intermediate values $\alpha$ & intermediate behavior & \olivieraccv{$\leq \min(\max(\alpha,1/d) bd,d) $}\\
    \bottomrule
    \end{tabular}
    \medskip
    \caption{Optimization algorithms used in the present paper. \baptisteaccv{The last setting is new compared to \cite{danglehre}. We modified the maximum mutation rate} $\alpha$ for doing a local or global search depending on $\alpha$, so that the diversity of the outputs is maintained when $\alpha$ is small (Sect. \ref{div}). $||x||_0$ denotes the number of non-zero components of $x$.}
    \label{tab}
\end{table}
\subsection{Open source codes}
We use the GAN publicly available at \url{https://github.com/moxiegushi/pokeGAN}, which is an implementation of Wasserstein GAN \cite{wgan}, the StyleGAN2 \cite{stylegan2} available at \url{thispersondoesnotexist.com}, and \baptisteaccv{PGAN on FashionGen from Pytorch GAN zoo \cite{pytorchganzoo}}.
Koncept512 is available at \url{https://github.com/subpic/koniq}.
Our combination of Koncept512 and \baptisteaccv{PGAN} is available at DOUBLEBLIND.  %\url{https://github.com/facebookresearch/pytorch_GAN_zoo/tree/frugan_koncept_only}. 
We use the evolutionary programming platform Nevergrad~\cite{nevergrad}. 
%The additional datasets are Mountains, Clouds, Flags. They are built using images found on internet (details in Table \ref{xptable}). 
\section{Experiments}
%Sections 3.1 - 3.3 present applications of EvolGAN to StyleGAN2 (datasets: Faces (FFHQ), cats, horses and artworks), PokeGAN (datasets: mountains and Pokemons) and PytorchGanZoo (dataset: FashionGen) respectively.
\baptisteaccv{We present applications of \frugan{} on three different GAN models:
(i) StyleGAN2 for faces, cats, horses and artworks (ii) PokeGAN for mountains and Pokemons (iii) PGAN from Pytorch GAN zoo for FashionGen.}
\subsection{Quality improvement on StyleGAN2}
\begin{figure}[t]
    \centering
    \includegraphics[width=.49\textwidth]{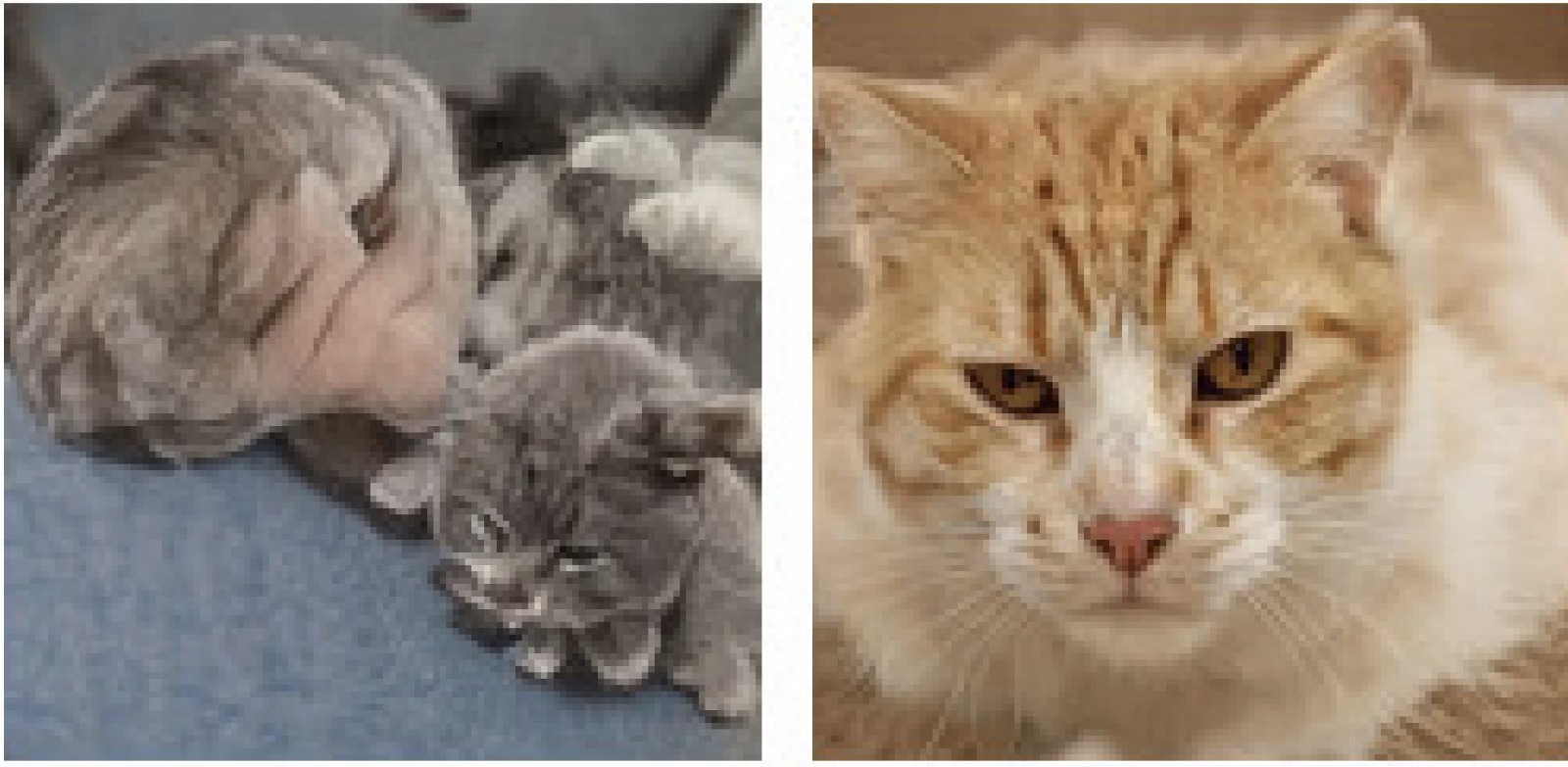}
    \includegraphics[width=.49\textwidth]{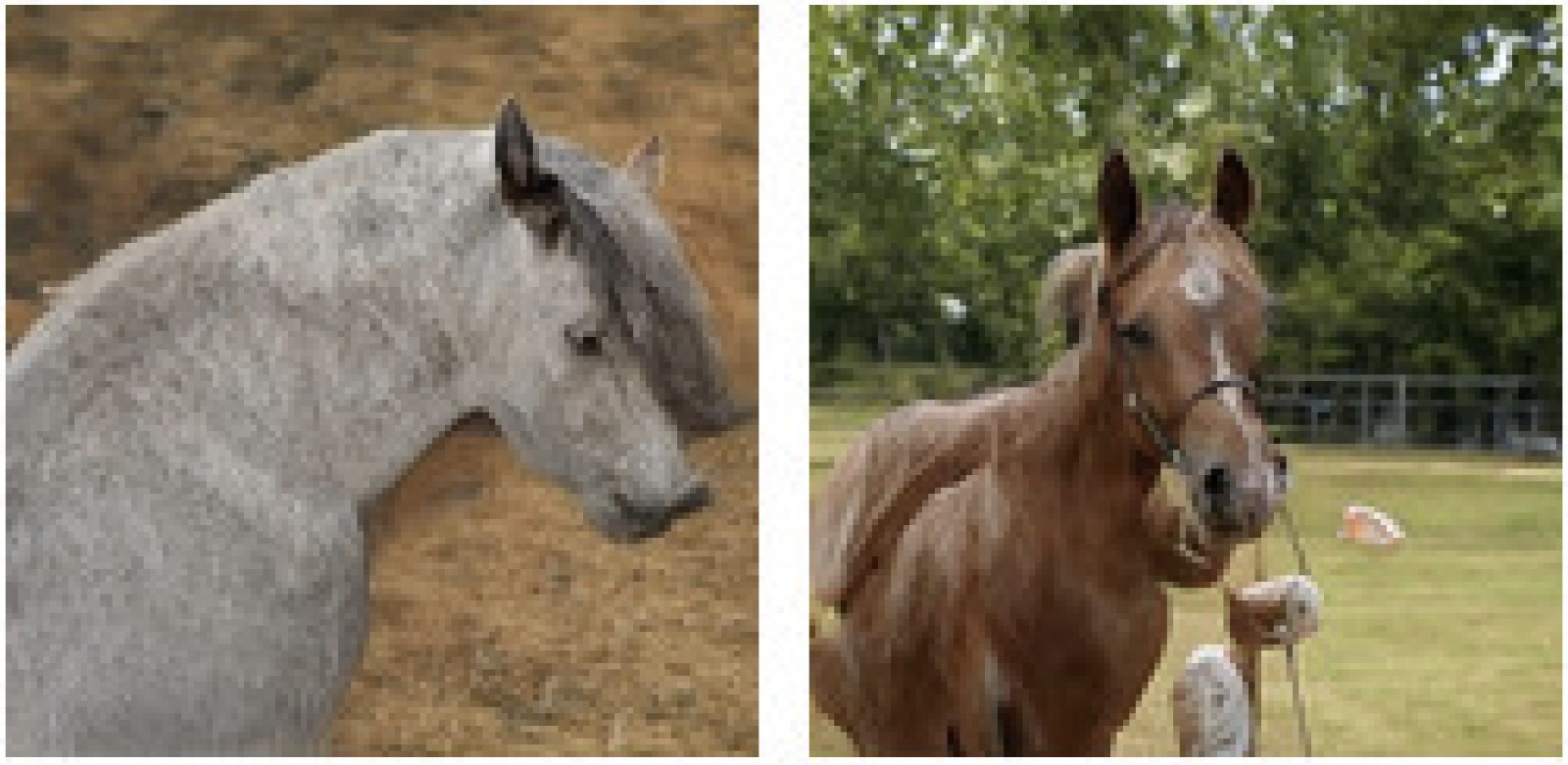}\\
     (a) Cherry-picked~~~~~~~~~(b) Random ~~~~~~~~~(c) Cherry-picked \hfill (d) Cherry-picked \hfill ~ \\
    ~~~~~poor StyleGAN2~~~~~~~~~~~~\frugan{} \hfill~~~~~~~~~~~~ poor \frugan{} \hfill ~~~~~~~ poor StyleGAN2\hfill ~ \\
    ~~~~~generation \hfill ~~~~~~~~~~~generation \hfill ~~~~~generation \hfill ~~~~~generation \hfill ~ \\
    \caption{For illustration, bad generations by StyleGAN2 and by EvolGAN. (a) Generation of a cat by StyleGAN2: we looked for a bad generation and found that one. Such bad cases completely disappear in the \frugan{} counterpart. (b) example of cat generation by \frugan{}: we failed to find a really bad \frugan{} generated image. (c) Example of bad horse generation by \frugan{}: the shape is unusual (looks like the muzzle of a pork) but we still recognize a horse. (d) Bad generation of a horse by StyleGAN2: some hair balls are randomly flying in the air.}
    \label{badcasts}
\end{figure}

\baptisteaccv{\brthree{The} experiments are based on open source codes \cite{nevergrad,pytorchganzoo,pokegan,koncept512reference,ava}.} We use the StyleGAN2~\cite{stylegan2} trained on a horse \fabienaaccv{dataset, a face dataset, an artwork dataset, and a cat dataset\footnote{\url{https://www.thishorsedoesnotexist.com/},\url{https://www.thispersondoesnotexist.com/},\url{https://www.thisartworkdoesnotexist.com/},\url{https://www.thiscatdoesnotexist.com/}}}. 
%We use the styleGAN2~\cite{stylegan2} trained on a horse dataset, a face dataset, an artwork dataset, and a cat dataset\footnote{\br{The results of t}hese GANs are visible at \url{https://www.thispersondoesnotexist.com/}, \url{https://www.thiscatdoesnotexist.com/} \ft{, \url{https://thisartworkdoesnotexist.com/}} and \url{https://www.thishorsedoesnotexist.com/}.}.
\olivieraccv{{\bf{Faces.}}}
\ccc{We conducted a human study to assess the quality of EvolGAN compared to StyleGAN, by asking to 10 subject their preferred generations \olivieraccv{(pairwise comparisons, double-blind, random positioning)}. \brcameraready{There were 11 human raters in total, including both experts with a strong photography background and beginners. 70\% of the ratings came from experts.} Results appear in Table \ref{ava}.} 
Faces are the most famous result of StyleGAN2. 
\baptisteaccv{Although the results are positive as the images generated by \frugan{} are preferred significantly more than 50\% of the time, the difference between StyleGAN2 and \frugan{} is quite small on this essentially solved problem compared to wild photos of cats or horses or small datasets.}
%  Table \ref{toto} present LPIPS results, evaluating the diversity\cite{lpips,zhu2017toward} of FashionGen generations. The metric is computed on 1000 images for PGAN and \frugan{} variants.} \ccc{Higher values correspond to higher diversity of samples.}
%\brthree{The} results \brthree{are of} excellent quality, and therefore our method \brthree{fails to lead to substantial improvements in average}.
\begin{table}[t]
    \centering
    \begin{tabular}{ccccc}
     %   \multicolumn{5}{c}{Quality estimator $q=\koncept{}$.}\\
            \toprule
            & $\frugan{}_{1, \infty}=G$   & $\frugan{}_{10,\infty}$   &  $\frugan{}_{20, \infty}$ &  $\frugan{}_{40,\infty}$    \\
            \midrule
        $\frugan{}_{10,\infty}$  & 60.0 & &     &     \\
        $\frugan{}_{20,\infty}$  & 50.0 & 57.1 &     &     \\
        $\frugan{}_{40,\infty}$  & 75.0 & 44.4 & 66.7  &     \\
        $\frugan{}_{80,\infty}$  & 53.8 & 53.8 & 40.0  &    46.2 \\ 
	    10-80  ag-& & & & \\
	    gregated  & 60.4\% $\pm$ 3.4\% (208 ratings) & & & \\
        %$\frugan{}_{500,\infty}$ & 54.0$\pm$ 5.0\% &  & & \\ % 24/50, more data to come
        \bottomrule
    \end{tabular}
    \medskip
    \caption{\baptisteaccv{Human study on faces dataset.} $\alpha=\infty$, quality estimator $q=\koncept{}$. Row $X$, col. $Y$: frequency at which human raters preferred $\frugan{}_{X,\infty}$ to $\frugan{}_{Y,\infty}$. By construction, for all $\alpha$, $\frugan{}_{1,\alpha}$ is equal to the original GAN. The fifth row aggregates all results of the first four rows for more significance.  }%\olivieraccv{Numbers are above $50\%$: using \frugan{} for modifying the latent vector $z$ improves the original StyleGAN2, but not by much.}}% $G$ (by construction equal to $\frugan{}_{1, \infty}$).}
    \label{k}
    \label{ava}
\end{table}
\begin{table}[t]
    \centering
    %Results with \koncept{}\\
    \begin{tabular}{cccc}
    \toprule
    Dataset & Budget $b$ & Quality estimator & score  \\
    \midrule
	    Cats & 300 & \koncept{} & 83.71 $\pm$1.75\% (446 ratings)\\ %()
    %Cats & 300 & \koncept{} & 16/22+25/30+15/20+37/40+45/49+12/20+17/21+203/240 +1/2+6/8= 138/161=TODO85.7\% $\pm$ 2.8\%  TODO  \\
	    Horses & 300 & \koncept{} & 70.43 $\pm$ 4.27\% (115 ratings)\\
	    Artworks & 300 & \koncept{} & 69.19 $\pm$ 3.09\% (224 ratings) \\   % (53+21+23+12+46) (75+27+30+17+75)
    %Horses & 300 & \koncept{} & 22/27 + 10/18+35/50=67/95 +14/20=TODO70.5\%$\pm$ 4.7\% TODO \\
    \bottomrule
    \end{tabular}
    \medskip
    \caption{Difficult test beds. $\alpha=\infty$; same protocol as in Tables \ref{k} i.e. we check with which probability human raters prefer an image generated by $\frugan{}_{G,b,\alpha}$ to an image generated by the original Gan $G$. By definition of \frugan{}, $\forall \alpha,G=\frugan{}_{G,1,\alpha}$. Number are above $50\%$: using \frugan{} for modifying the latent vector $z$ improves the original StyleGAN2.}
    \label{kanimals}
\end{table}
\begin{table}[t]
    \centering
    \begin{tabular}{ll}
    \toprule
Context & LPIPS score \\
 \midrule
%   \multicolumn{2}{c}{\textbf{FashionGen}}\\
 PGAN & 0.306 $\pm$ 0.0003\\
$\frugan{}_{PGAN,b=40,\alpha=0}$& 0.303 $\pm$ 0.0003\\
$\frugan{}_{PGAN,b=40,\alpha=1}$ & 0.286 $\pm$ 0.0003\\
$\frugan{}_{PGAN,b=40,\alpha=\infty}$ & 0.283 $\pm$ 0.0002\\
%   \multicolumn{2}{c}{\textbf{Artworks: low ratio budget/dimension preserves diversity}}\\
% StyleGan2 & 0.781 $\pm$ 0.003\\
% $\frugan{}_{StyleGan2,b=10,\alpha=1}$ & 0.795 $\pm$ 0.007\\
% $\frugan{}_{StyleGan2,b=40,\alpha=1}$ &0.797 +/- 0.014  \\
% $\frugan{}_{StyleGan2,b=80,\alpha=1}$ & 0.795 +/- 0.02 \\
%For the artworks: 0.79534 +/- 0.00728 for onetenth, 0.78052 +/- 0.00269 for others and 0.80328 +/- 0.00886 for the random mix from onetenth and others with size of of onetenth
\bottomrule
\end{tabular}
\smallskip
    \caption{\olivieraccv{LPIPS scores on FashionGen. As expected, $\alpha=0$ mostly preserves the diversity of the generated images, while higher values of $\alpha$ can lead to less diversity for the output of \frugan{}.} \baptisteaccv{The LPIPS was computed on samples of $50,000$ images for each setting.} }%Artworks: computed on $147$ images: no significant differences; slight improvement actually, maybe by reducing blurred/average images.}%TODOTODO
    \label{toto}
\end{table}
\olivieraccv{\bf{Harder settings.}}
Animals and artworks are a much more difficult setting (Fig. \ref{badcasts}) - StyleGAN2 sometimes fails to propose a high quality image. Fig. \ref{badcasts} presents examples of generations of $StyleGAN2$ and $\frugan{}_{StyleGAN2,b,\alpha}$ in such cases. Here, $\frugan{}$ has more headroom for providing improvements \br{than for} faces: results are presented in Table \ref{kanimals}. The case of horses or cats is particularly interesting: the failed generations often contain globally \br{unrealistic} elements, \br{such as} random hair balls flying in the air or unusual positioning of limbs, which \baptisteaccv{are} removed by \frugan{}. 
\olivieraccv{\ccc{For illustration purpose, in Fig. \ref{badcasts}} we present a few examples of generations which go wrong for the original StyleGan2 and for $\frugan{}_{StyleGan2,b=100,\alpha=0}$; the bad examples in the case of the original StyleGan2 are much worse. } 
%it is not clear how \koncept{} can diagnose such problems.
\subsection{Small \baptisteaccv{difficult} datasets \olivieraccv{ and $\alpha=0$}}\label{quality}
\begin{figure}[ht]
    \centering
    \includegraphics[width=.4\textwidth]{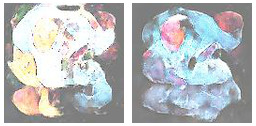}\ \ \ \ \
    \includegraphics[width=.4\textwidth]{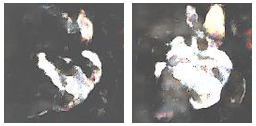}
    \caption{In both cases, a Pokemon generated by the default GAN (left) and after improvement by \koncept{} (right). For the left pair, after improvement, we see eyes and ears for a small elephant-style pokemon sitting on his back. A similar transformation appears for the more rabbit-style pokemon on the right hand side. These cherry-picked examples \olivieraccv{(cherry-picked, i.e. we selected specific cases for illustration purpose)} are, however, less convincing than the randomly generated examples in Fig. \ref{pairingpgan} - Pokemons are the least successful applications, as \koncept{}, with $\alpha$ large or big budgets, tends to push those artificial images towards dark almost uniform images.}
    \label{mignon}
\end{figure}

%We consider quality improvement and diversity preservation  with $\alpha$ small for very small datasets such as PokeGAN.
% \todo{this paragraph is hard to understand}
\br{In this section we focus on the use of \frugan{} for  small datasets.}
We use the original pokemon dataset in PokeGAN~\cite{pokegan} \br{and} an additional dataset created from copyright-free images of mountains.
The previous section was quite successful, using $\alpha=\infty$ \br{(i.e. random search)}. % (a.k.a a single epoch evolutionary algorithm). 
The drawback is that the obtained images are not necessarily related to the original ones, and we might lose diversity \olivieraccv{(though Section \ref{div} shows that this is not always the case, see discussion later)}. We will see that $\alpha=\infty$ fails in the present case. 
\br{In this section, we use} $\alpha$ small, and check if the obtained images $\frugan{}_{G,b,\alpha,z}$ are better than $G(z_0)$ (see Table \ref{xptable}) and close to the original image $G(z_0)$ (see Fig.\ref{mignon}).
Fig. \ref{mignon} presents a Pokemon generated by the default GAN and its improved counterpart obtained by $\frugan{}$ with $\alpha=0$.
%{\bf{Quality improvement.}}
Table \ref{xptable} presents our experimental setting \br{and the results of our human study conducted} on PokeGAN. We see a significant improvement when using \koncept{} {\textit{on real-world data}} (as opposed to drawings such as Pokemons, for which \koncept{} fails), whereas we fail with \ava{} as in previous experiments (see Table \ref{ava}). We succeed on drawings with \koncept{} only with $\alpha=0$: on this dataset of drawings (poorly adapted to \koncept{}), $\alpha$ large leads to a pure black image.
\begin{table}[t]\scriptsize\centering
    \begin{tabular}{|c|c|c|c|c|c|c|}
        \hline
        %Type of & Number of & Number of & budget & Quality & $\alpha$ & Frequency \\
        %images & images & training epochs & $b$ & estimator & & of  image \\
        % & & & & & preferred \\
        %& & & & & & to original \\
        Type & Number of & Number of & Budget  & Quality & $\alpha$ & Frequency \\
        of & of & training  & $b$  & estimator & & of  image \\
        images & images & epochs &               & & & preferred to original\\
        \hline
        \multicolumn{7}{|c|}{Real world scenes}\\
        \hline
	    Mountains & 84 & 4900 & 500 &  \koncept{} & $0$ & 73.3\% $\pm$ 4.5\% (98 ratings)\\
        \hline
%        Clouds    & 249 & TODO & 500 & \ava{}     & $0$ &  \\
%        Clouds    & 249 & TODO & 500 & \koncept{} & $0$ &  \\
%        \hline
        \multicolumn{7}{|c|}{Artificial scenes}\\
%        \hline
%        Flags     & 196 & 4900 & TODO& & & \\
        \hline
        Pokemons  & 1840 & 4900 & 500 & \koncept{} & $0$& 55\% \\
        Pokemons  & 1840 & 4900 & 2000 & \koncept{} & $0$& 52\% \\        
	    Pokemons  & 1840 & 4900 & 6000 & \koncept{} & $0$& 56.3 $\pm$ 5.2\% (92 ratings) \\    
%        Pokemons  & 1840 & 4900 & (500, 2000, 6000 aggregated) & \koncept{} & $0$ & 55\% $\pm$ 2.5\% \\
        \hline
        \multicolumn{7}{|c|}{Artificial scenes, higher mutation rates}\\
        \hline
        Pokemons  & 1840 & 4900 & 500 & \koncept{} & $1/7$& 36.8\% \\
        Pokemons  & 1840 & 4900 & 20 &        \koncept{} & $\infty$ & 0\% \\
        \hline
    \end{tabular}\medskip
    \caption{\label{xptable}Experimental results with $\frugan{}_{PokeGAN,b,\alpha=0}$. Reading guide: the last column shows the probability that an image $\frugan{}_{PokeGAN,b,\alpha=0,z}=G(z^*(z_0))$ was prefered to the starting point $PokeGAN(z_0)$. The dimension of the latent space is $d=256$ except for mountains ($d=100$). \koncept{} \brthree{performs well} on real world scenes but not on artificial scenes. % - 
    \brthree{For Pokemon with $\alpha = \infty$, the 0\% (\fabienaaccv{0 success} out of 24 tests!) is  interesting:} the code starts to generate almost uniform images even \brthree{with a budget $b=20$,} showing that \koncept{} fails on drawings. On mountains (the same GAN, but trained on real world images instead of Pokemons), and \brthree{to a lesser extent} on Pokemons for small $\alpha$, \brthree{the images generated using \frugan{} are preferred more than $50\%$ of the time:} using \frugan{} for modifying the latent vector $z$ improves the original PokeGan network.}
\end{table}
%{\bf{Robustness improvement:}} TODO
%\end{itemize}
%For this we generate several pairs $(z_i,z^*(z_i))$ and check that a human can easily match the pairs in a big pool with such pairs shuffled all together (Fig. \ref{pairing}), at least for small $\alpha$.
\def\nopk{
\begin{figure}[t]
    \centering
    \includegraphics[width=.98\textwidth]{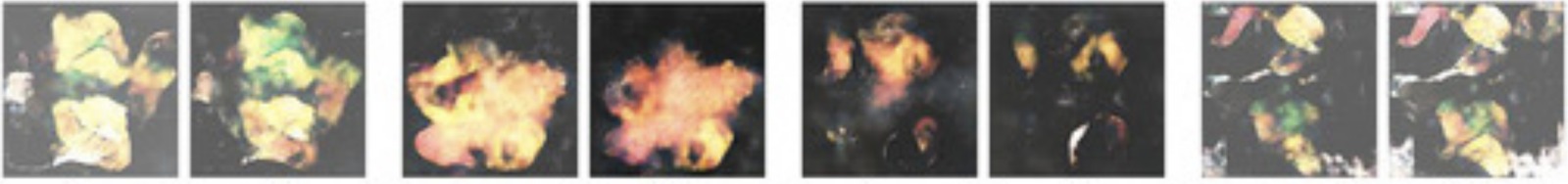}
    \medskip   \caption{\olivieraccv{Pokemons: a case with synthetic images, in which \koncept{} is not relevant.} Images before optimization ($G(\baptisteaccv{z_0})$ for randomly drawn $\baptisteaccv{z_0}$) and after optimization ($\frugan{}_{G,b,\alpha=0,z}=G(z^*(\baptisteaccv{z_0}))$), with $G=PokeGAN$. We also see that when \brthree{optimizing the generation of drawings}, we tend to lose parts of the image when pushing the optimization of the \koncept{} score too far (see in particular the $3^{rd}$ pair). \brthree{There is no such issue} on real world images. This also explains why $\alpha=0$ is better than $\alpha>0$ on Pokemons (Table \ref{xptable}), though \brthree{the} results remain disappointing (there is \brthree{some} improvement, but not much). \brthree{On the other hand, \frugan{} is always effective} on real world scenes (Table \ref{kanimals}). See also Fig. \ref{pairingpgan}.}
    \label{pairing}
\end{figure}}
\subsection{Quality improvement}\label{pgz}
\begin{figure}
    \centering
    \begin{tabular}{ccc|ccc}
         $G(z_0)$ & $G(z^*_{40}(z_0))$ & $G(z^*_{320}(z_0))$ & $G(z_0)$ & $G(z^*_{40}(z_0))$ & $G(z^*_{320}(z_0))$  \\
       \includegraphics[width=.16\textwidth]{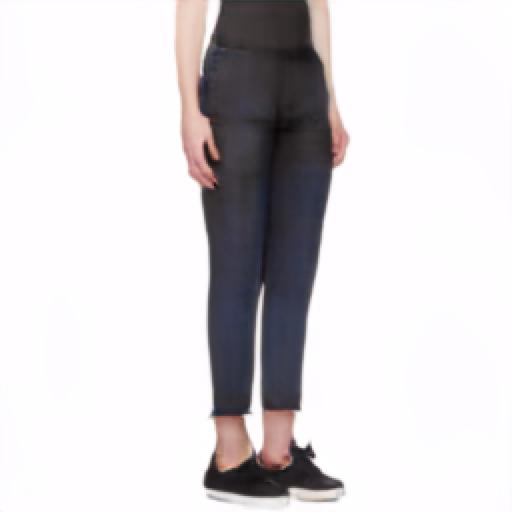} & \includegraphics[width=.16\textwidth]{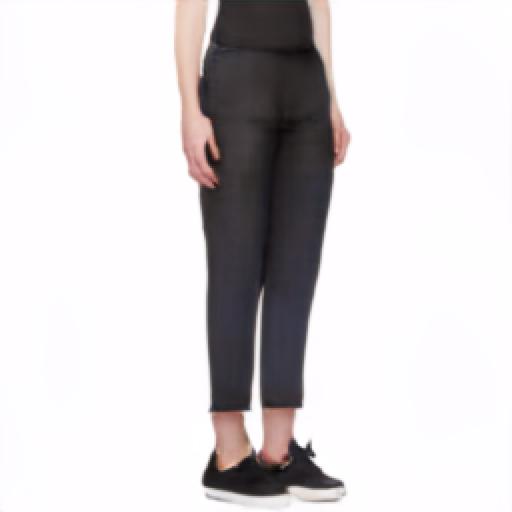} & \includegraphics[width=.16\textwidth]{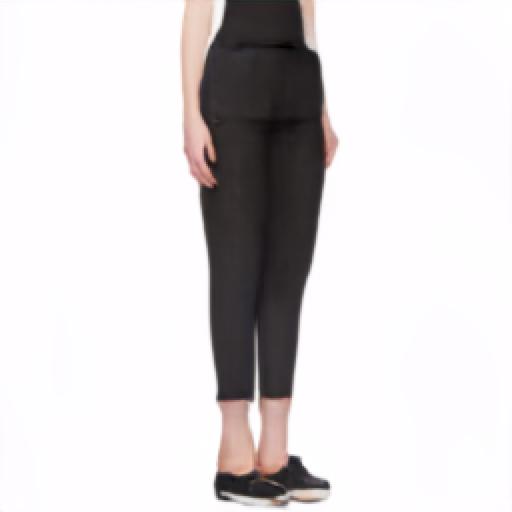}
        & \includegraphics[width=.16\textwidth]{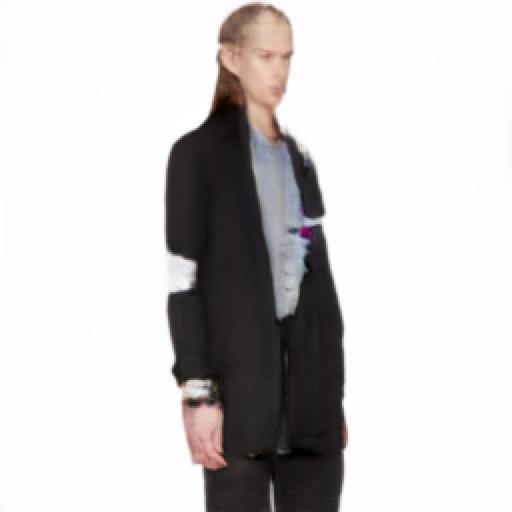} & \includegraphics[width=.16\textwidth]{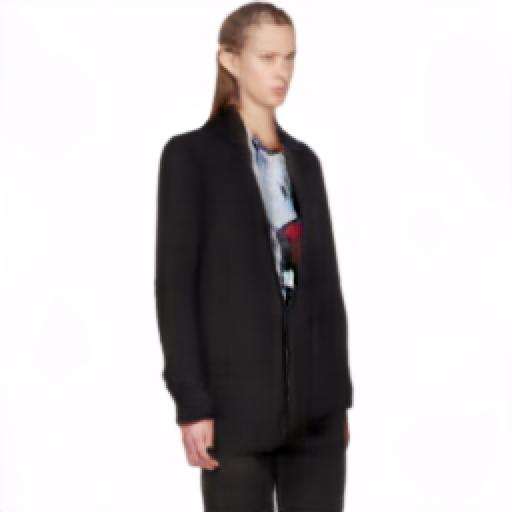} & \includegraphics[width=.16\textwidth]{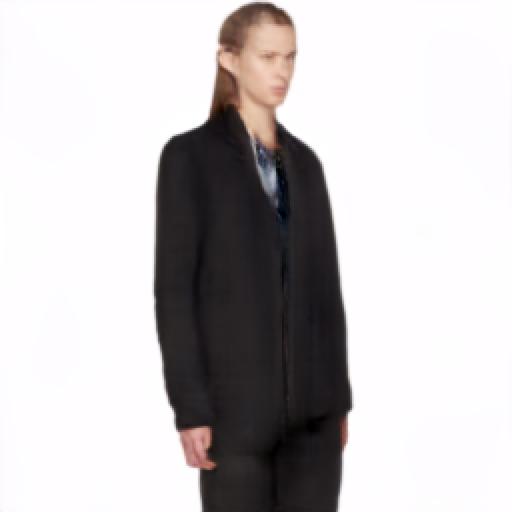}\\
        \includegraphics[width=.16\textwidth]{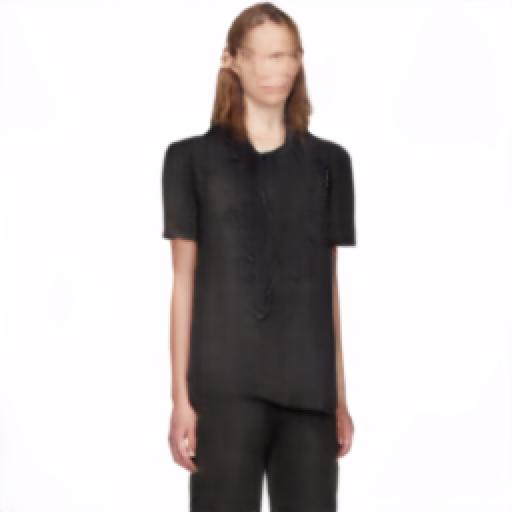} & \includegraphics[width=.16\textwidth]{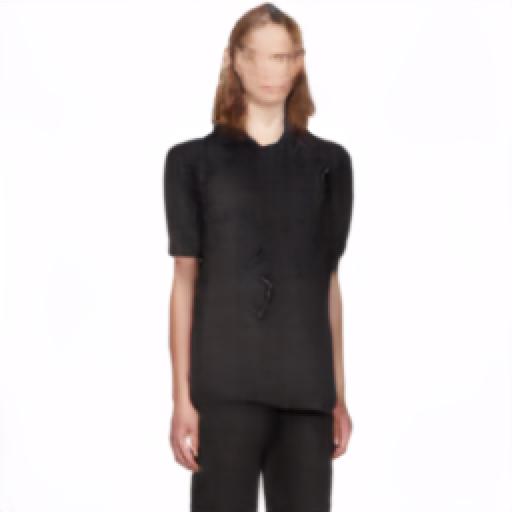} & \includegraphics[width=.16\textwidth]{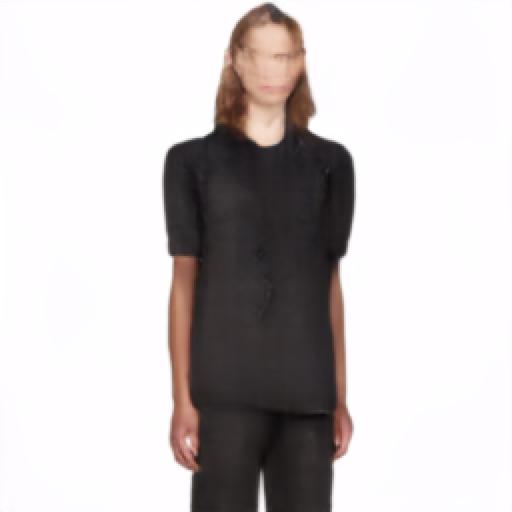}
        & \includegraphics[width=.16\textwidth]{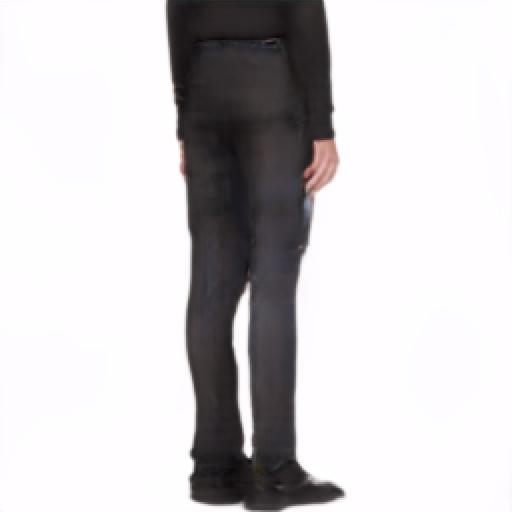} & \includegraphics[width=.16\textwidth]{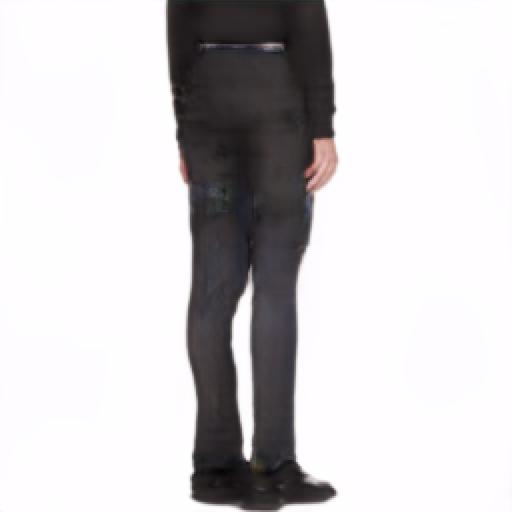} & \includegraphics[width=.16\textwidth]{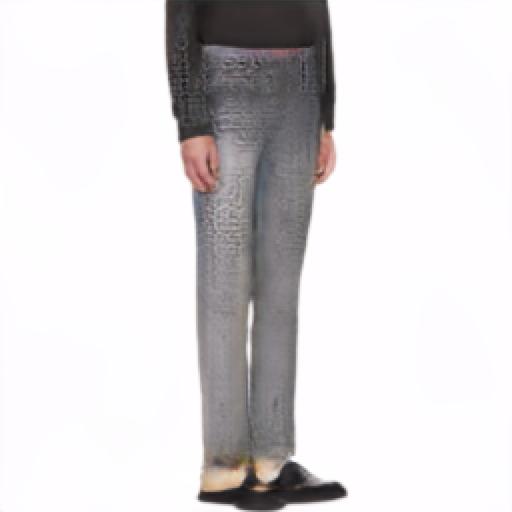}\\
        \includegraphics[width=.16\textwidth]{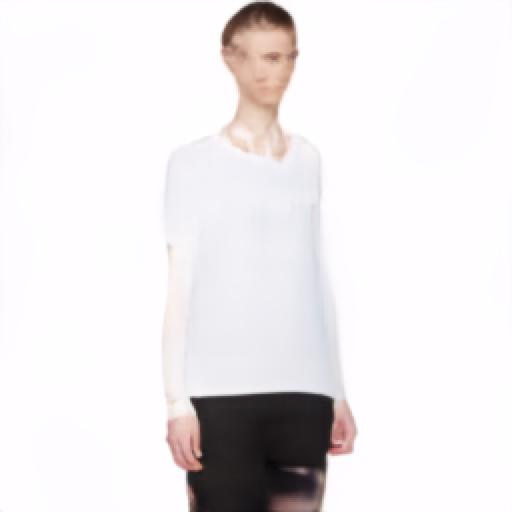} & \includegraphics[width=.16\textwidth]{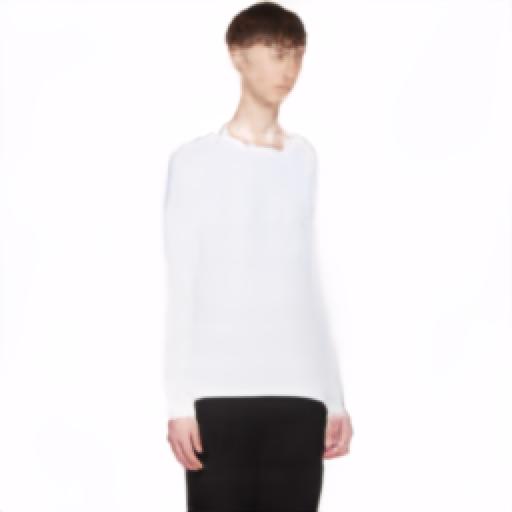} & \includegraphics[width=.16\textwidth]{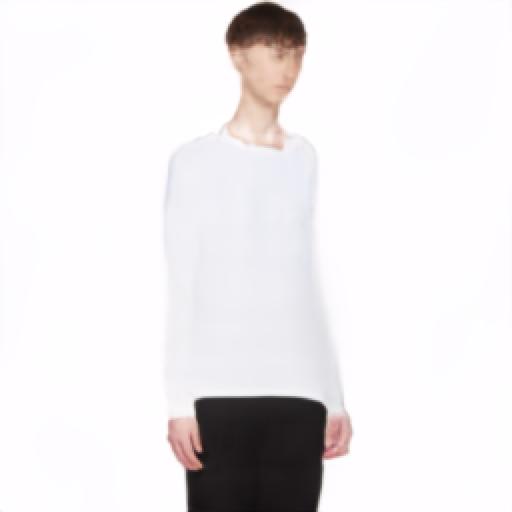}
        & \includegraphics[width=.16\textwidth]{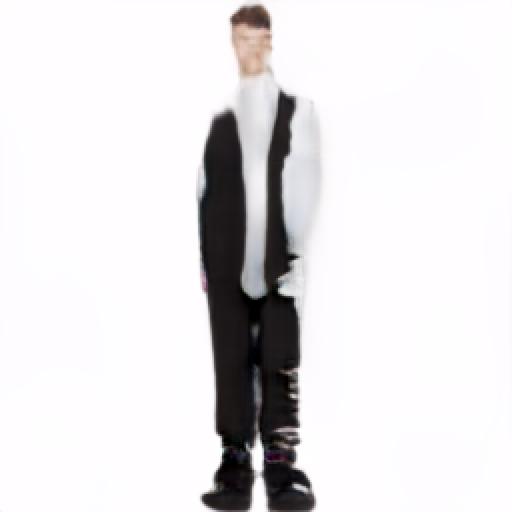} & \includegraphics[width=.16\textwidth]{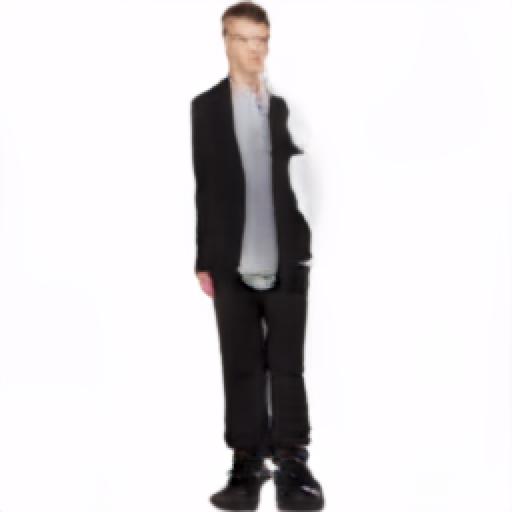} & \includegraphics[width=.16\textwidth]{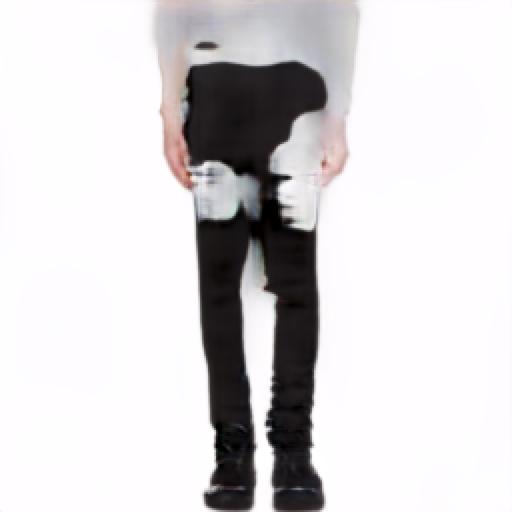}\\
        \includegraphics[width=.16\textwidth]{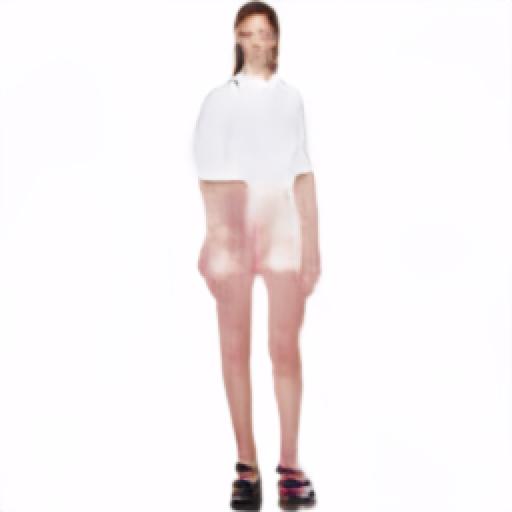} & \includegraphics[width=.16\textwidth]{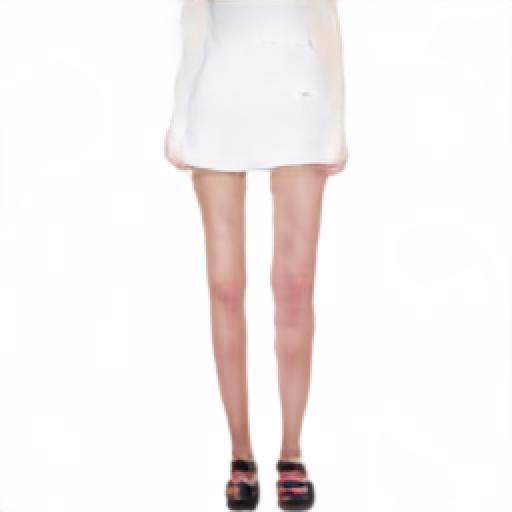} & \includegraphics[width=.16\textwidth]{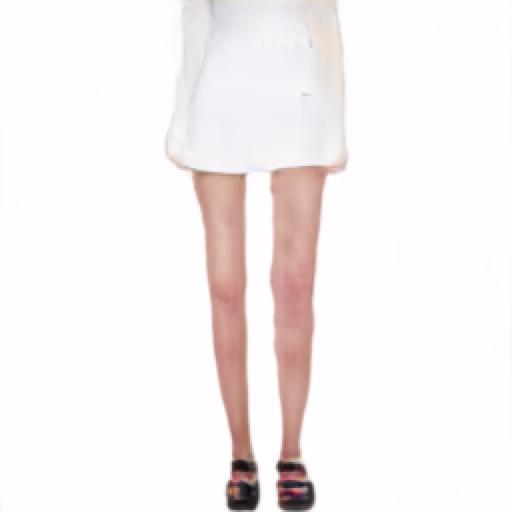}
        & \includegraphics[width=.16\textwidth]{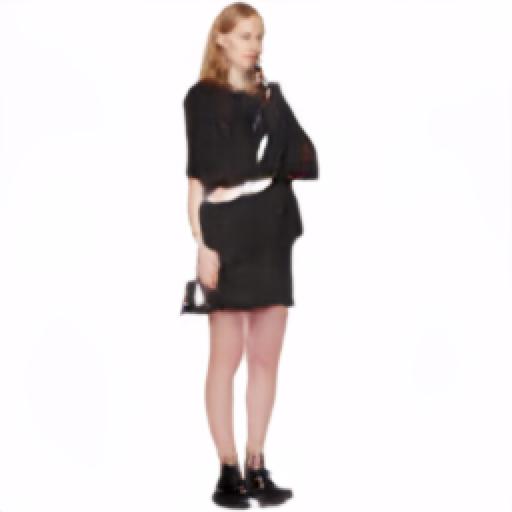} & \includegraphics[width=.16\textwidth]{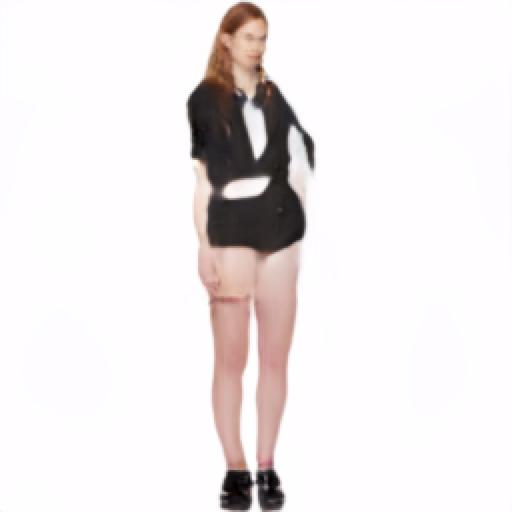} & \includegraphics[width=.16\textwidth]{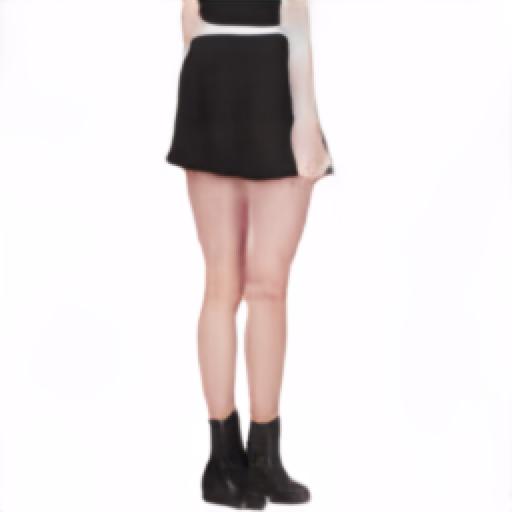}\\
    \end{tabular}
    %\includegraphics[trim = {0 0 0 0}, clip, width=.47\textwidth]{pairing/fashion_pairing_40_DiscreteOnePlusOne.jpg}\\
     %   \includegraphics[trim = {0 0 0 0}, clip, width=.47\textwidth]{pairing/fashion_pairing_320_DiscreteOnePlusOne.jpg}\\
        %~\hfill Budget $b=40$ \hfill Budget $b=320$ \hfill ~\\
    \caption{\olivieraccv{Preservation of diversity, in particular with budget $40$, when $\alpha=0$. We present} triplets $PGAN(z),PGAN(z^*_{40}(z))=\frugan{}(PGAN,b=40,\alpha=0,z),PGAN(z^*_{320}(z))=\frugan{}(PGAN,b=320,\alpha=0,z)$, i.e. in each case the output of PGAN and its optimized counterpart with budgets 40 and 320 respectively (images 1, 4, 7, 10, 13, 16, 19, 22 are the $G(z)$ and images 2, 5, 8, 11, 14, 17, 20, 23 are their counterparts $G(z^*_{40}(z))$ for budget $40$; indices +1 for budget 320). With $\alpha=1$ (\fabienaaccv{unpresented}) $PGAN(z)$ and $\frugan{}_{PGAN,b,\alpha,z}$ are quite different, so that we can not guarantee that diversity is preserved. With $\alpha=0$ diversity is preserved with $b=40$: for each of these 8 cases, the second image ($b=40$) is quite close to the original image, just technically better~---~except the $8^{th}$ one for which $G(z)$ is quite buggy and \frugan{} can rightly move farther from the original image. Diversity is less preserved with $b=320$: e.g. on the top right we see that the dress becomes shorter at $b=320$.}
    \label{pairingpgan}
\end{figure}

Pytorch Gan Zoo \cite{pytorchganzoo} is an implementation of progressive GANs (PGAN\cite{karras2017progressive}), applied here with FashionGen~\cite{fashionGen2018} as a dataset. \br{The dimension of the latent space is 256.}
In Table \ref{pganquality}, we present the results of our human study comparing $\frugan{}_{PGAN,b,\alpha}$ to $\frugan{}_{PGAN,1,\alpha}=PGAN$. \br{With $\alpha=0$, humans prefer \frugan{} to the baseline in more than 73\% of cases, even after only 40 iterations. $\alpha=0$ also ensures that the images stay close to the original images when the budget is low enough (see Table \ref{tab}).}
 Fig. \ref{pairingpgan} shows some examples of generations using \frugan{} and the original PGAN.%: $G(z^*(\baptisteaccv{z_0}))$ is very similar to $G(\baptisteaccv{z_0})$, so the diversity is preserved.  % the dimension of the latent space is 256, and we use the released checkpoints.
\begin{table}[t]
    \begin{center}
 \begin{tabular}{|c|c|}
    \hline
        b & Frequency \\
        \hline
        \multicolumn{2}{|c|}{{\bf{$\alpha=0$}}}\\
        \hline
	    \bf 40  & \bf 73.33 $\pm$ 8.21\% (30 ratings)) \\   % (14+8) (20+2+8)
	    \bf 320 & \bf 75.00 $\pm$ 8.33\% (28 ratings)) \\   % (14+7) (18+3+7)
	    \bf 40 and 320 aggreg. & \bf 74.13 $\pm$ 5.79\% (58 ratings))\\   % (14+8+14+7) (20+2+8+18+3+7)
        \hline
        \multicolumn{2}{|c|}{$\alpha=1$} \\        
        \hline
	    40 & 48.27 $\pm$ 9.44\% (29 ratings)\\   % (11+3) (18+8+3)
	    320 & 67.74 $\pm$ 8.53\% (31 ratings)\\   % (16+5) (20+6+5)
	    40 and 320 aggreg. & 58.33 $\pm$ 6.41\% (60 ratings)\\   % (11+3+16+5) (18+8+3+20+6+5)
        \hline
%        \end{tabular}
%        \end{center}
%
%        \begin{center}
%        \begin{tabular}{|c|c|}
%    \hline
%        b & Frequency \\
%        \hline
        \multicolumn{2}{|c|}{$\alpha=\infty$}  \\
        \hline
		40  &  56.66 $\pm$ 9.20\% (30 ratings)\\   % (14+3) (21+6+3)
		320 &  66.66 $\pm$ 9.24\% (27 ratings) \\   % (14+4) (17+6+4)
		40 and 320 aggreg. & 61.40 $\pm$ 6.50\% (57 ratings)\\   % (14+14+3+4) (21+6+3+17+6+4)
        \hline
        \multicolumn{2}{|c|}{All $\alpha$ aggregated}\\
        \hline
		40  & 59.55 $\pm$ 5.23\% (89 ratings)\\   % (14+11+14+8+3+3) (21+18+20+2+8+8+3+6+3)
		320 & 69.76 $\pm$ 4.98\% (86 ratings)\\   % (14+16+14+7+5+4) (18+20+17+3+7+6+5+6+4)
		40 and 320 aggreg. &    64.57 $\pm$ 3.62\% (175 ratings)\\   % (14+11+14+8+3+3+14+16+14+7+5+4) (21+18+20+2+8+8+3+6+3+18+20+17+3+7+6+5+6+4)
        \hline
    \end{tabular}
        \end{center}
    % \medskip
    \caption{Frequency (a.k.a score) at which various versions of $\frugan{}_{PGAN,b,\alpha,z}=PGAN(z^*(\baptisteaccv{z_0}))$ were preferred to their starting point $PGAN(z)$ on the FashionGen dataset.
    \otc{This experiment is performed with \koncept{} as a quality \br{estimator}.} In most experiments we get the best results with $\alpha=0$ and do not need more than a budget $b=40$. \brthree{The values are} greater than $50\%$, meaning that \frugan{} improves the original PGAN network on FashionGen \brthree{according to human preferences.}}
    \label{pganquality}
\end{table}
{\subsection{Consistency: preservation of diversity.}}\label{div}
 %In the present section we show that the obtained images have a satisfactory diversity. 
\vladaccv{Here} we show that the \vladaccv{generated image} is close to the one before the optimization. More precisely, given $z\mapsto G(z)$, the \vladaccv{following two} methods provide related outputs:
 \vladaccv{method 1 (classical GAN) outputs $G(\baptisteaccv{z_0})$, and method 2 (\frugan{}) outputs $\frugan{}_{G,b,\alpha,z}=G(z^*\baptisteaccv{(z_0)}\fabienaaccv{)}$}, where $z^*(\baptisteaccv{z_0})$ is obtained by our evolutionary algorithms starting at $z$ with budget $b$ and parameter $\alpha$ (Sect. \ref{local}).
 \olivieraccv{Fig. \ref{pairingpgan} shows some \vladaccv{example generated images} using PGAN and \frugan{}. For most examples, $G(z^*(z_0))$ is very similar to $G(z_0)$ so the diversity of the original GAN is preserved \vladaccv{well}.} 
 % In addition, Table \ref{toto} shows numerically \baptisteaccv{that the diversity of the images generated using \frugan{} is similar to that of the original GAN,} in particular with $\alpha=0$.}
 \baptisteaccv{Following \cite{lpips,zhu2017toward}, we measure numerically the diversity of the \vladaccv{generated images} from the PGAN model, and \olivieraccv{from} \frugan{} models based on it, using the LPIPS score. The scores were computed on samples of $50,000$ images generated with each method. For each sample, we computed the LPIPS with another randomly-chosen generated image. The results are presented in Table~\ref{toto}. \ccc{Higher values correspond to higher diversity of samples.} \frugan{} preserves the diversity of the images generated when used with $\alpha=0$.}
 % or when the ratio budget/dimension is small.}
\subsection{Using \ava{} rather than \koncept{}}
% In Table \ref{allava} we show that \ava{} looks less convenient for \frugan{} than \koncept{}.
\br{In Table \ref{allava} we show that \ava{} is less suited than \koncept ~ as a quality assessor in \frugan{}.
The human annotators do not find the images generated using \frugan{} with \ava{} to be better than those generated without \frugan{}.
We hypothesize that this is due to the subjectivity of what \ava{} estimates: aesthetic quality. 
While humans generally agree on the factors accounting for the technical quality of an image (sharp edges, absence of blur, right brightness), annotators often disagree on aesthetic quality.
Another factor may be that aesthetics are inherently harder to evaluate than technical quality.}
\begin{table}[t]
    \centering
    \begin{scriptsize}
     \begin{tabular}{ccccc}
     \toprule
            %\multicolumn{5}{c}{Quality estimator $q=\ava{}$.}\\
            & $\frugan{}_{1, \infty}=G$   & $\frugan{}_{10,\infty}$   &  $\frugan{}_{20, \infty}$ &  $\frugan{}_{40,\infty}$    \\
            \midrule
        $\frugan{}_{10,\infty}$  & 34.8&      &     &        \\
        $\frugan{}_{20,\infty}$  & 52.0 & 42.8 &     &       \\
        $\frugan{}_{40,\infty}$  & 39.1& 32.0 & 36.4&        \\
        $\frugan{}_{80, \infty}$ & 52.2& 52.2 & 40.9& 56.0\% \\
    $\frugan{}_{10-80,\infty}$ & 44.5\% $\pm$ 5.0\%  & & &   \\
     (aggregated) & & & & \\
    %500 & 22/40 + 11/29+103/200 TODO  & & & \\
    500 & 50.55 $\pm$ 3.05 \% & & & \\
    \bottomrule
    \end{tabular}\\
    (a) Faces with StyleGAN2: reproducing Table \ref{k} with \ava{} in lieu of \koncept{}.\\
    %\medskip
    %Results with \ava{}\\
    \medskip
    \begin{tabular}{cccc}
    \toprule
    Dataset & Budget $b$ & Quality estimator & score  \\
%        &          & estimator &        \\
    \midrule
    %Cats & 300 & \ava{} & 11/20+13/31+TODO  \\
    Cats & 300 & \ava{} & 47.05 $\pm$7.05\%  \\
    Artworks & 300 & \ava{} & 55.71 $\pm$ 5.97 \% \\ % (39/70)
    \bottomrule
    \end{tabular}\\
    (b) Reproducing Table \ref{kanimals} with \ava{} in lieu of \koncept{}.\\
        \medskip
    \begin{tabular}{ccccccc}
    \toprule
        Type & Number of & Number of & Budget  & Quality & $\alpha$ & Frequency \\
        of & of & training  & $b$  & estimator & & of  image \\
        images & images & epochs &               & & & preferred to original\\
%        & & & & & & to original \\
        \midrule
        Mountains & 84 & 4900 & 500 &  \ava{} & $0$ & 42.5\% \\
        Pokemons  & 1840 & 4900 & 500 & \ava & $0$& 52.6\% \\
        Pokemons  & 1840 & 4900 & 500 & \ava & $1/13$ & 52.6\% \\
        \bottomrule
    \end{tabular}\\
    (c) Reproducing Table \ref{xptable} with \ava{} rather than \koncept{}.\\
\end{scriptsize}
        \medskip
    \caption{\br{Testing \ava{} rather than \koncept{} as a quality estimator. With \ava{}, \frugan{} fails to beat the baseline according to human annotators.}}
    \label{allava}
\end{table}
\section{Conclusion}
We have shown that, given a generative model $z\mapsto G(z)$, optimizing $z$ by an evolutionary algorithm using \koncept{} as a criterion and preferably with $\alpha=0$ (i.e. the classical $(1+1)$\fabienaaccv{-Evolution Strategy}), leads to
\begin{itemize}
    \item \baptisteaccv{The generated images are preferred by humans as shown on} Table \ref{kanimals} for StyleGAN2, Table \ref{xptable} for PokeGan and Table \ref{pganquality} \baptisteaccv{for PGAN on FashionGen}
    \item \baptisteaccv{The diversity is }preserved, \baptisteaccv{as shown on} Fig. \ref{pairingpgan} and Table~\ref{toto}, when using a small value for $\alpha$ (see Section \ref{local}).
\end{itemize}

{\bf{Choosing $\alpha$.}}
$\alpha$ small, i.e., the classical $(1+1)$\fabienaaccv{-Evolution Strategy} with mutation rate $1/d$, is usually the best choice: we preserve the diversity (with provably a small number of mutated variables, and experimentally a  resulting image differing from the original one mainly in terms of quality), and the improvement compared to the original GAN is clearer as we can directly compare $\frugan{}_{G,b=d,\alpha=0,z}$ to $G(z)$~---~a budget $b\simeq d/5$ was usually enough. Importantly, evolutionary algorithms clearly outperformed random search and not only in terms of speed: we completely lose the diversity with random search, as well as the ability to improve a given point. Our application of evolution is a case in which we provably preserve diversity~---~with a mutation rate bounded by $\max(\alpha,1/d)$, and a budget $b=d/5$, and a dimension $d$, we get an expected ratio of mutated variables at most $b \times \max(\alpha,1/d)$. In our setting $b=40,d=256,\alpha=0$ \brthree{so the maximum expected ratio of mutated variables} is at most $40/256$ in Fig. \ref{pairingpgan}. A tighter, run-dependent bound, can be obtained by comparing $\baptisteaccv{z_0}$ and $\baptisteaccv{z^*(z_0)}$ and could be a stopping criterion.

{\bf{Successes.}} We get an improved GAN without modifying the training. \brthree{The} results are positive in all cases in particular difficult real-world data (Table \ref{kanimals}), though the gap is moderate when the original model is already excellent (faces, Table \ref{k}) or when data are not real-world (Pokemons, Table \ref{xptable}.% - see also discussion in Fig. \ref{pairing}).
\br{\frugan{} with \koncept{} is particularly successful} on several difficult cases with real-world data\,\textemdash\,Mountains with Pokegan, Cats, Horses and Artworks with StyleGAN2 and FashionGen with Pytorch Gan Zoo. %\,\textemdash\,. %i.e. all real-world images except Faces for which there is not much headroom. 
%The only clear failures are for artificial images, and \ava{}, which is not adapted to our work.

{\bf{Remark on quality assessement.}}
\koncept{} can be used on a wide range of applications. As far as our framework can tell, it outperforms \ava{} as a tool for \frugan{} (Table \ref{allava}). However, it fails on artificial scenes such as Pokemons, unless, we use a small $\alpha$ for staying local.

{\bf{Computational cost.}}
All the experiments with PokeGAN \br{presented here} could be run on a laptop without using any GPU.
\br{The e}xperiments with StyleGAN2 and PGAN use at most 500 (and often just 40) calls to the original GAN, without any specific retraining: we just repeat the inference with various latent vectors $z$ chosen by our evolutionary algorithm as detailed in Section \ref{zefrugan}.
%Experiments with StyleGAN2 are performed on that same laptop but with an access to the internet interface.
% Experiments include Pokemons, Artworks, Horses, Cats, FashionGen, Mountains and Faces: only for faces, we failed to get positive results. % already said above 
%\subsection*{Further work}

% \subsection*{Acknowledgements} We are grateful to C. Couprie for suggestions in early phases of this work and for proofreading the final manuscript.

\section*{Acknowledgments} 
Funded by the Deutsche Forschungsgemeinschaft (DFG, German Research Foundation), Project-ID 251654672, TRR 161 (Project A05). 

%\clearpage
\FloatBarrier
\bibliographystyle{splncs}
\bibliography{camille,lsca,doe,bibliodoe,bibliolsca,biblio}%,biblio2}
\end{document}